\pdfoutput=1

\documentclass[11pt]{article}
\usepackage[]{latex/acl}

\usepackage{soul}
\usepackage{times}
\usepackage{multirow}
\usepackage{latexsym}
\usepackage{soul}
\usepackage{comment}

\usepackage[T1]{fontenc}

\usepackage[utf8]{inputenc}

\usepackage{microtype}

\usepackage{inconsolata}

\usepackage{amsmath,amsfonts,bm}

\def\eqref#1{equation~\ref{#1}}

\def\1{\bm{1}}

\DeclareMathAlphabet{\mathsfit}{\encodingdefault}{\sfdefault}{m}{sl}
\SetMathAlphabet{\mathsfit}{bold}{\encodingdefault}{\sfdefault}{bx}{n}

\DeclareMathOperator*{\argmin}{arg\,min}

\usepackage{graphicx}
\usepackage{diagbox}
\usepackage{adjustbox}
\usepackage{pgfplots}
\usepackage{subfigure}
\usepackage{booktabs} 

\usepackage{url}

\usepackage{breakurl}
\usepackage[breaklinks]{hyperref}

\definecolor{darkblue}{rgb}{0.0,0.0,0.65}
\definecolor{bettergreen}{rgb}{0.0,0.7,0.0}
\definecolor{Gray}{gray}{0.85}

\DeclareUnicodeCharacter{2060}{\nolinebreak}

\hypersetup{
  colorlinks = true,
  citecolor  = darkblue,
  linkcolor  = darkblue,
  citecolor  = darkblue,
  filecolor  = darkblue,
  urlcolor   = darkblue,
}

\usepackage{subfiles}
\usepackage{url}
\usepackage{caption}
\usepackage{graphicx}
\usepackage{amsfonts}
\usepackage{amsmath}
\usepackage{amsthm}
\usepackage{subfigure}
\usepackage{amssymb}
\usepackage{mathtools}
\usepackage{bbm}
\usepackage{algpseudocode}
\usepackage{algorithm}
\usepackage{enumitem}
\usepackage{amssymb}
\usepackage{bbm}
\usepackage{wrapfig}

\theoremstyle{definition}

\newtheorem{nullhypothesis}{Null Hypothesis}
\newcommand{\Sref}[1]{\S\ref{#1}}

\newcommand{\LM}{P_{\text{LM}}}
\newcommand{\embd}{M_{\text{embd}}}
\newcommand{\dcos}{d_{\cos}}
\newcommand{\method}{\textsc{SemStamp}}
\newcommand{\methodk}{$k$-\method\ }

\newcommand{\shortmethod}{\textsc{SStamp}}
\newcommand{\shortmethodk}{$k$-\shortmethod}

\newcommand{\secvsabove}{\vspace{-1mm}}
\newcommand{\secvsbelow}{\vspace{-1mm}}
\newcommand{\subsecvs}{\vspace{-1mm}}

\newcommand{\figvsbottom}{\vspace{-3mm}}
\newcommand{\paravs}{\vspace{-1.5mm}}
\newcommand{\eqvs}{\vspace{-1.5mm}}

\newlength\myindent
\title{\method: A Paraphrase-Robust  Watermark}
\title{\method: A Watermark Robust to Semantic Paraphrases}
\title{Optimizing Paraphrasing Robustness for Natural Language Watermark}
\title{A Semantic Watermark Algorithm for Language Model Generation with Robustness to Paraphrase Attacks}
\title{\method: Paraphrase-Robust Semantic Watermark}
\title{\method: A Paraphrase-Robust Semantic Watermark}
\title{\method: Semantic Watermarking with Paraphrase Invariance}
\title{\method: A Semantic Watermark}
\title{\method: Semantically Watermarking Text Generation with Paraphrastic Robustness}
\title{\method: Semantic Watermarking \\ with  Paraphrastic Robustness}
\title{\method: Paraphrase-Robust Watermarked \\ 
Text Generation}
\title{\method: Semantically Watermarking Text via Paraphrastic Robustness}
\title{\method: Semantically Watermarking Language Model Texts with Paraphrastic Robustness}
\title{\method: Semantic Watermarking Text Generation with Paraphrastic Robustness}
\title{\method: Semantic Watermarked Generation \\ with Paraphrastic Robustness} 
\title{A Simple Yet Effective Variant of \method \ for Machine-Generated Text Detection}
\title{\methodk: A Simple Yet Effective Variant of Semantic Watermark for Machine-Generated Text Detection}
\title{
\vspace*{-0.5in}
{{\small \hfill ACL-Findings'24}\\
\vspace*{.25in}} 
\methodk: A Clustering-Based Semantic Watermark for\\ Detection of Machine-Generated Text}

\author{
Abe Bohan Hou\textsuperscript{$\clubsuit$} \quad 
Jingyu Zhang\textsuperscript{$\clubsuit$} \quad 
Yichen Wang\textsuperscript{$\diamondsuit$} \\
\textbf{Daniel Khashabi}\textsuperscript{$\clubsuit$} \quad 
\textbf{Tianxing He}\textsuperscript{$\heartsuit$} \\ 
\textsuperscript{$\clubsuit$}Johns Hopkins University \quad \textsuperscript{$\heartsuit$}University of Washington \quad \textsuperscript{$\diamondsuit$}{Xi'an Jiaotong University}\\
\texttt{\{bhou4, jzhan237\}@jhu.edu\quad goosehe@cs.washington.edu}
}

\begin{document}

\maketitle

\begin{abstract}
Recent watermarked generation algorithms inject detectable signatures during language generation to facilitate post-hoc detection. While token-level watermarks are vulnerable to paraphrase attacks, \method\ \citep{hou23semstamp} applies watermark on the semantic representation of sentences and demonstrates promising robustness. \method \ employs locality-sensitive hashing (LSH) to partition the semantic space with arbitrary hyperplanes, which 
may lead to
a suboptimal trade-off between robustness and speed. We propose $k$-\method, a simple yet effective enhancement of \method, utilizing $k$-means clustering as an alternative of LSH to partition the embedding space with awareness of inherent semantic structure. Experimental results indicate that \methodk saliently improve its robustness and sampling efficiency while preserving the generation quality, advancing a more effective tool for machine-generated text detection.
\end{abstract}

\secvsabove
\section{Introduction}
\secvsbelow
\label{sec:intro}

\begin{figure}
    \begin{center}
    \vspace{-2mm}
    \includegraphics[width=0.48\textwidth]{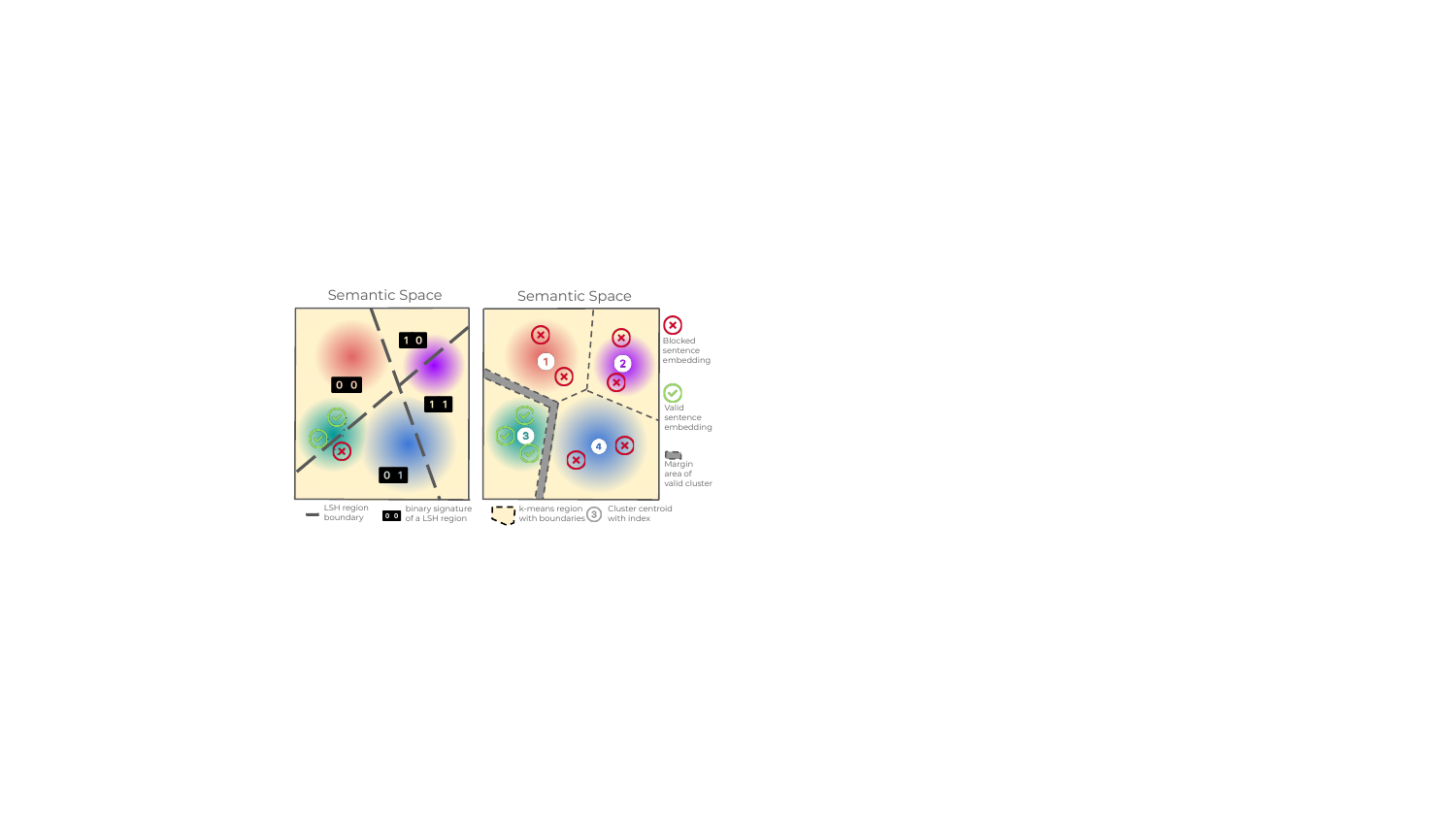}   
    \vspace{-6mm}
    \end{center}
  \caption{Illustrations of the semantic space. Sentence embeddings with close meanings share similar colors. \textbf{(Left)} Random planes from LSH arbitrarily partition the semantic space and split similar sentences into different regions. \textbf{(Right)} Margin-based rejection in $k$-\method. Sentence embeddings which fall into the gray-shaded areas of a valid region will be rejected.} 
  \figvsbottom
  \label{fig:lshkvisual}
\end{figure}

To facilitate the detection of machine-generated text \citep{mitchell-10.1145/3287560.3287596}, recent watermarked generation algorithms usually inject detectable signatures \citep[\textit{i.a.}]{Kuditipudi2023RobustDW, Yoo2023RobustMN, Wang2023TowardsCT, Christ2023UndetectableWF, Fu2023WatermarkingCT, hou23semstamp}. A major concern for these approaches is their robustness to potential attacks, since a malicious user could attempt to remove the watermark with text perturbations such as editing and paraphrasing \citep{Wang2024StumblingBS, krishna2023paraphrasing, sadasivan2023aigenerated, kirchenbauer2023reliability, zhao2023provable}. \citet{hou23semstamp} propose \method, a paraphrase-robust and sentence-level watermark which assigns signatures to each watermarked sentence according to the locality sensitive hashing (LSH) \citep{indyk98lsh} partitioning of semantic space (see \ref{subsec:prelim}). While demonstrating promising robustness against paraphrase attacks, \method\ arbitrarily partitions the semantic space by a set of \emph{random} hyperplanes, possibly splitting semantically similar sentences into different partitions (see Fig.\ref{fig:lshkvisual}).

This limitation motivates our proposed method, \methodk (detailed in \Sref{subsec:sem-k}), which partitions the space 
via
$k$-means clustering \citep{lloyd1982kmeans} on the semantic structure of a given text domain (e.g. news, narratives, etc.). In \S\ref{sec:exp}, we show that the clustering-based partitioning in \methodk greatly improves its robustness against sentence-level paraphrase attacks and sampling efficiency.\footnote{
We have released   \href{https://github.com/bohanhou14/SemStamp}{the code} for reproducibility. 
 Corresponding authors: Abe Hou, Jingyu Zhang, and Tianxing He.}
 

\begin{figure*}[ht]
    \centering
    \includegraphics[scale=0.95]{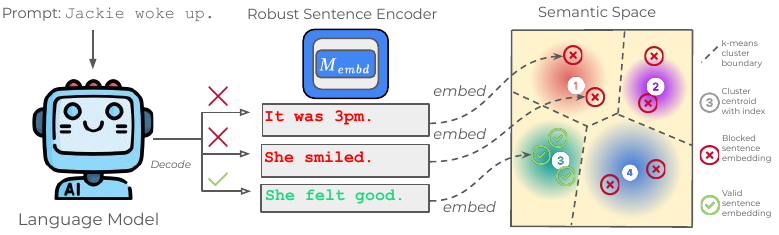}
    \caption{An overview of the proposed \methodk algorithm. $k$-means clustering partitions the semantic space into semantically similar regions. The sentence generation is accepted if the closest cluster of its sentence embedding corresponds to a "valid" region in the semantic space.}
    \figvsbottom
    \label{fig:teaser}
\end{figure*}
\begin{figure}
    \fbox{
    \begin{minipage}{18.5em}
    \footnotesize
    \textbf{Prompt: }
    In Chapter 18, Richard begins at Kenge and Carboy's.
    
     \textbf{Non-Watermarked Generation: } He goes to the inn where Mr. Kenge has been let off by the landlord. There, he meets a woman named Hannah, who is looking for him. He asks her where he is wanted. 
     
    \leavevmode\newline
    \textbf{\shortmethod: }
    He meets up with Lydgate, who is there to see if the money from the deal is still there. The lawyers are ready to go to trial, but Richard says he has a better plan. He wants to leave Middlemarch for good. 
    
    \leavevmode\newline
    \textbf{\shortmethodk:} He also sees Adam for the first time since his imprisonment. They discuss the latest updates in their respective personal lives. Adam is living with Dinah and is still angry with Adam for having to leave him. 
    \end{minipage}
    }
    \caption{Generation Examples of \methodk compared with \method.
    \textbf{Both generations are contextually sensible and coherent as compared to non-watermarked generations.}
    Additional examples after paraphrase are presented in Figure \ref{fig:textparaphraseexamples} in the Appendix. }
    \vspace{-3mm}
    \label{fig:textexamples}
\end{figure}
\begin{algorithm}[t]
\small
\caption{\methodk text generation algorithm and subroutines}
\label{alg:main-k}
\begin{algorithmic} 
\State \textbf{Input:} language model $\LM$, prompt $s^{(0)}$, the text domain $\mathcal{D}$,  the number of sentences to generate $T$.

\State \textbf{Params:} sentence embedding model fine-tuned on $\mathcal{D}$, $\embd^{\mathcal{D}}$ with embedding dimension $h$, maxout number $N_\text{max}$,
margin $m>0$, valid region ratio $\gamma \in(0,1)$, the number of $k$-means clusters $K$, a large prime number $p$, an integer $N$.
\State \textbf{Output:} generated sequence $s^{(1)}\dots s^{(T)}$.
\item[]

\Procedure{\methodk}{}
\State $C_K \gets \textsc{Initialize}(\mathcal{D}, K)$ to initialize $K$ cluster centroids based on $\mathcal{D}$.

\For{$t=1,2,\dots, T$}
 \begin{enumerate}[leftmargin=16.5mm,align=left]
\item Find the index of the closest cluster centroid of the previously generated sentence, $q^{(t-1)} \gets \textsc{Assign}(s^{(t-1)}, C_K)$, and use $q^{(t-1)} \cdot p$ as the seed to randomly divide the \textit{index set} of clusters $C_K$ into a ``valid region set'' $G^{(t)}$ of size $\gamma \cdot K$ and a ``blocked region set'' $R^{(t)}$ of size $(1-\gamma) \cdot K$.
\item {\textbf{repeat} Sample a new sentence from LM,     
    
    \textbf{until} the index of the closest cluster centroid of the new sentence, $q^{(t)}$, is in the ``valid region set'', and the margin requirement \textsc{Margin}($s^{(t)}, m$) is satisfied \textbf{or} sampling has repeated over $N_\text{max}$ times.

\item Append the selected sentence $s^{(t)}$ to context.
}\end{enumerate}
\EndFor
\State \textbf{return} $s^{(1)}\dots s^{(T)}$
\EndProcedure
\end{algorithmic}
\begin{algorithmic}
\item[] 
\Function{Initialize}{$\mathcal{D}, K$}
    \State $\mathcal{D}^{'}_{N} \sim \mathcal{D}$ // sample $N$ sentences from $D$
    \State $C_K \gets \textsc{k-means}(\mathcal{D}^{'}_{N}, K)$ // obtain $k$ cluster centroids
    \State \textbf{return} $C_K$
\EndFunction
\item[] 
\Function{Assign}{$s, C_K$} 

\hspace{-5mm}// find the index of the closest centroid by cosine distance
    \State \textbf{return} $\argmin_{i=1,\dots,K}\dcos(v, c_i)$, where $c_i \in C_K$~ 

\EndFunction
\end{algorithmic}
\end{algorithm}
\begin{algorithm}[t]
    \small
    \caption{\methodk detection algorithm}
    \label{alg:detect}
    \begin{algorithmic}
        \State \textbf{Input:} a piece of text $T$, saved $k$-means cluster centroids $C_K$
        \State \textbf{Params:} sentence embedding model finetuned on $\mathcal{D}$, $\embd^{\mathcal{D}}$, $z$-threshold range $Z$, human-written texts $H$, a large prime number $p$, valid region ratio $\gamma\in(0,1)$, number of $k$-means clusters $K$.
        \State \textbf{Output:} a $z$-score based on the ratio of detected sentences.
        
        \Procedure{Detect}{$T, C_K$}
        
        \State $s_1, ..., s_N \gets \textsc{Sentence-Tokenize(T)}$
        
        \State $q^{(1)} \gets \textsc{Assign}(s_1, C_K)$
        
        \State $\texttt{seed} \gets q^{(1)}\cdot p$
        
        \State $G^{(1)} \gets \textsc{Random-Sample}(\texttt{seed}, K, \gamma)$~ // pseudo-randomly sample a set of cluster centroid indices of size $K \cdot \gamma$, where the randomness of sampling is controlled by $\texttt{seed}$.
        \For{$t=2,\dots, N$} 
        \State $q^{(t)} \gets \textsc{Assign}(s_t, C_K)$
        
        \If {$q^{(t)} \in G^{(t-1)}$}
            \State $S_V$ += 1
        \EndIf
        
        $\textsc{seed} \gets q^{(t)}\cdot p$
        
        $G^{(t)} \gets \textsc{Random-Sample}(\texttt{seed}, K, \gamma)$
        \EndFor
        \EndProcedure
        \State $z \gets \frac{S_V - \gamma N}{\sqrt{\gamma (1-\gamma) N}}$ 
        \State \textbf{return} $z$ 
    \end{algorithmic}%
\end{algorithm}

\section{Approach}
\label{sec:approach}

We first review the  existing watermark algorithms for machine-generated text detection (\S\ref{subsec:prelim}) and introduce our proposed watermark (\S\ref{subsec:sem-k}).

\subsection{Preliminaries}
\label{subsec:prelim}
\paragraph{Token-Level Watermark}
\citet{kirchenbauer2023watermark} develop a notable token-level watermark algorithm. Given a token history $w_{1:t-1}$, the vocabulary $V$ is pseudo-randomly divided into a ``green list'' $G^{(t)}$ and a ``red list'' $R^{(t)}$, where a hash of the previous token $w_{t-1}$ is used as the seed of the partition. The algorithm then adds a bias to the logits of all tokens in the green-list and sample the next token with an increased probability from the green-list. For a given piece of text, the watermark can be detected by conducting one proportion $z$-test (detailed in \Sref{app:z-score}) on the number of green list tokens. 
\vspace{-6mm}
\paragraph{\method}
\label{subsec:semantic_wm}
Under the intuition that common sentence-level paraphrase modifies tokens but preserves sentence meaning,  \citet{hou23semstamp} introduce \method \ to apply watermark on sentence semantics by partitioning the embedding space with locality sensitive hashing (LSH). 

To initialize the LSH partitioning, $d$ normal vectors are randomly sampled from a Gaussian distribution to specify $d$ hyperplanes in the semantic space $\mathbb{R}^h$. For an embedding vector $v\in\mathbb{R}^h$, a $d$-bit binary LSH signature is assigned, where each digit specifies the position of $v$ in relation to each hyperplane. Each signature $c\in\{0,1\}^d$ indexes a region consisting of all vectors with signature $c$. 

During generation, given a sentence history denoted by $s^{(0)}\dots s^{(t-1)}$, the space of signatures is pseudorandomly partitioned into a set of ``valid'' regions $G^{(t)}$ and a set of ``blocked'' region $R^{(t)}$.  The LSH signature of the last generated sentence
is used as the random seed to control randomness. A new sentence generation, $s^{(t)}$, will be accepted and if its embedding belongs to any valid region, and rejected otherwise. To detect the watermark in a given piece of text, a one-proportion $z$-test is performed on the number of sentences whose signatures belong to valid regions (see \Sref{app:z-score}).
%


\subsection{\methodk}
\subsecvs
\label{subsec:sem-k}

As discussed earlier, \method{}   partitions the semantic space with \emph{random} planes, which could potentially  separate semantically similar sentences into two different regions, as shown in Fig.\ref{fig:lshkvisual}. Paraphrasing sentences near the margins of regions may shift their sentence embeddings to a nearby region, resulting in suboptimal watermark strength. This weakness motivates our proposed $k$-\method, a simple yet effective enhancement of \method \ that partitions the semantic space with $k$-means clustering \citep{lloyd1982kmeans}. 



To initialize \methodk, we assume the language model generates text in a specific domain $\mathcal{D}$ (e.g., news articles, scientific articles, etc.). We aim to model the semantic structure of $\mathcal{D}$ and partition its semantic space into $k$ regions. Concretely, we first randomly sample a large number of data from $\mathcal{D}$. We obtain their sentence embeddings with a robust sentence encoder fine-tuned on $\mathcal{D}$ with contrastive learning (detailed in \Sref{app:finetune}). We cluster the sentence embeddings into $K$ clusters with $k$-means \citep{lloyd1982kmeans} and save the cluster centroids. We index a region with $i \in \{1, ..., K\}$ representing the set of all vectors assigned to the $i$-th centroid.

The generation process is analogous to \method\ \citep{hou23semstamp}, as illustrated in Fig.\ref{fig:teaser}: given a sentence history $s^{(0)}\dots s^{(t-1)}$, $K$ regions are pseudorandomly partitioned into a set of valid regions $G^{(t)}$ of size $\gamma \cdot K$ and a set of blocked regions $R^{(t)}$ of size $(1-\gamma) \cdot K$, where $\gamma \in (0, 1)$ is the ratio of valid regions. The cluster assignment of $s^{(t-1)}$, $C(s^{(t-1)})$, seeds the randomness of the partition at time step $t$, where $C(.)$ returns the cluster index by finding the closest cluster centroid of the input sentence embedding. We then conduct rejection sampling and only sentences whose embeddings fall into any valid regions (i.e., $C(s) \in G^{(t)} $) are accepted while the rest are rejected. If no valid sentence is accepted after a preset maxout number ($N_\text{max}$) of tries, the last decoded sentence will be chosen. The full algorithm is presented in Algo \ref{alg:main-k}.

\paravs
\paragraph{Cluster Margin Constraint}
To prevent the sampled sentences from being assigned to a nearby cluster after paraphrasing, we propose a cluster margin constraint similar to \citep{hou23semstamp}. We constrain the sentence embeddings to be sufficiently away from the cluster boundaries (visualized in Fig.\ref{fig:lshkvisual}). Concretely, the cosine distance ($\dcos$) of the candidate sentence embedding ($v$) to the closest centroid ($c_q$) needs to be smaller than other cluster centroids by at least a margin $m$:
\vspace{-2mm}
\begin{equation}
    \vspace{-1mm}
    \eqvs
     \dcos(v, c_q)  < \min_{i\in \{1,\dots,K\} \setminus q} \dcos(v, c_i) - m,
\end{equation}
where 
$q$ is the index of the closest cluster centroid to $v$, i.e.,
$q = \argmin_{i=1,\dots,K}\dcos(v, c_i)$, and $v = \embd(s^{(t)})$ is the embedding of the generated sentence at time step $t$ by a robust sentence embedder $\embd$.



The detection procedure of \methodk is analogous to \method{} which uses one-proportion $z$-test on the number of sentences belong to valid regions, explained in \Sref{app:z-score} and Algo~\ref{alg:detect}.

\secvsabove
\section{Experiments}
\secvsbelow
\label{sec:exp}

\subsection{Experimental Setup}
\subsecvs

Following \citet{hou23semstamp}, we conduct paraphrase attack experiments and compare the detection robustness of watermarked generations.

\paragraph{Task and Metrics}

We evaluate 1000 watermarked generations after paraphrase, respectively on the RealNews subset of the C4 dataset \citep{raffel2020exploring} and on the BookSum dataset \citep{kryscinski2021booksum}. We paraphrase watermarked generations sentence-by-sentence with the Pegasus paraphraser \citep{zhang2020pegasus}, Parrot used in \citet{sadasivan2023aigenerated}, and GPT-3.5-Turbo \citep{openai2022chatgpt}. We also implement the strong bigram paraphrase attack as detailed in \citet{hou23semstamp}. Detection robustness of paraphrased watermarked generations is measured with area under the receiver operating characteristic curve (\textbf{\textit{AUC}}) and the true positive rate when the false positive rate is at 1\% and 5\% (\textbf{\textit{TP@1\%}}, \textbf{\textit{TP@5\%}}).\footnote{ We denote machine-generated text as the ``positive'' class and human text as the ``negative'' class. A piece of text is classified as machine-generated when its $z$-score exceeds a threshold chosen based on a given false positive rate. See \Sref{app:z-score}. } Generation quality is measured with perplexity (\textbf{\textit{PPL}}) (using OPT-2.7B \citep{zhang2022opt}), trigram text entropy \citep{Zhang2018GeneratingIA} (\textbf{\textit{Ent-3}}), i.e., the entropy of the trigram frequency distribution of the generated text, and \textbf{\textit{Sem-Ent}} \citep{sement2022}, an automatic metric for semantic diversity. Following the setup in \citet{sement2022}, we perform $k$-means clustering ($k=50$) with the last hidden states of OPT-2.7B on text generations, and Sem-Ent is defined as the entropy of semantic cluster assignments of test generations.
We also measure the paraphrase quality with BERTScore \citep{zhang2019bertscore} between original generations and their paraphrases.

\begin{table*}[]
\centering
\scalebox{0.6}{
\begin{tabular}{@{}lllllllll@{}}
\toprule
\multicolumn{9}{c}{\textit{\textbf{AUC $\uparrow$}} / \textit{\textbf{TP@1\% $\uparrow$}} / \textit{\textbf{TP@5\% $\uparrow$}}} \\
~ Domain & Algorithm & No Paraphrase & Pegasus & Pegasus-bigram & Parrot & Parrot-bigram & GPT3.5 & GPT3.5-bigram \\ \midrule
\multicolumn{1}{c}{} & KGW & 99.6 / 98.4 / 98.9 & 95.9 / 82.1 / 91.0 & 92.1 / 42.7 / 72.9 & 88.5 / 31.5 / 55.4 & 83.0 / 15.0 / 39.9 & 82.8 / 17.4 / 46.7 & 75.1 / \ \ 5.9 / 26.3 \\

& SIR &99.9 / 99.4 / 99.9 & 94.4 / 79.2 / 85.4 & 94.1 / 72.6 / 82.6 &93.2 / 62.8 / 75.9 &95.2 / 66.4 / 80.2 & 80.2 / 24.7 / 42.7 & 77.7 / 20.9 / 36.4 \\

\multicolumn{1}{c}{} & \method & 99.2 / 93.9 / 97.1 & 97.8 / 83.7 / 92.0 & 96.5 / 76.7 / 86.8 & 93.3 / 56.2 / 75.5 & 93.1 / 54.4 / 74.0 & 83.3 / 33.9 / 52.9 & 82.2 / 31.3 / 48.7 \\

\multicolumn{1}{c}{\multirow{-4}{*}{\texttt{RealNews}}} & \methodk & 99.6 / 98.1 / 98.7 & \textbf{{99.5} / 92.7 / 96.5} & \textbf{99.0 / 88.4 / 94.3}& \textbf{97.8 / 78.7 / 89.4 }& \textbf{97.5 / 78.3 / 87.3} & \textbf{90.8 / 55.5 / 71.8} & \textbf{88.9 / 50.2 / 66.1} \\ \midrule

& KGW & 99.6 / 99.0 / 99.2 & 97.3 / 89.7 / 95.3 & 96.5 / 56.6 / 85.3 & 94.6 / 42.0 / 75.8 & 93.1 / 37.4 / 71.2 & 87.6 / 17.2 / 52.1 & 77.1 / \ \ 4.4  / 27.1 \\

& SIR & 1.0 / 99.8 / 1.0 & 93.1 / 79.3 / 85.9 & 93.7 / 69.9 / 81.5 & 96.5 / 72.9 / 85.1 & 97.2 / 76.5 / 88.0  & 80.9 / 39.9 / 23.6 &75.8 / 19.9 / 35.4 \\

 & \method & 99.6 / 98.3 / 98.8 & 99.0 / \textbf{94.3} / 97.0 & 98.6 / 90.6 / 95.5 & 98.3 / 83.0 / 91.5 & 98.4 / 85.7 / 92.5 & 89.6 / 45.6 / 62.4 & 86.2 / 37.4 / 53.8 \\

\multicolumn{1}{c}{\multirow{-4}{*}{\texttt{BookSum}}} & \methodk & 99.9 / 99.1 / 99.4 & \textbf{99.3} / 94.1 / \textbf{97.3} & \textbf{99.1 / 92.5 / 96.9 } & \textbf{98.4 / 86.3 / 93.9} & \textbf{98.8 / 88.9 / 94.9 }& \textbf{95.6 / 65.7 / 83.0} & \textbf{95.7 / 64.5 / 81.4} \\
\bottomrule
\end{tabular}
}
\caption{Detection results against various paraphrase attacks. All numbers in each cell are in percentages and correspond to \textbf{\textit{AUC}}, \textbf{\textit{TP@1\%}}, and \textbf{\textit{TP@5\%}}, respectively. All three metrics prefer higher values. KGW and SIR  refer to the watermarks in \citet{kirchenbauer2023watermark} and \citet{liu2023semantic}.
\textbf{\methodk\ is more robust than \method \ and KGW across most paraphrasers and their bigram attack variants and both datasets.}}
\label{tab:detection} 
\vspace{-2mm}
\end{table*}

\begin{table*}[t]
\centering
\scalebox{0.75}{\begin{tabular}{@{}lllllll@{}}
\toprule
 \multicolumn{7}{c}{\textit{\textbf{AUC $\uparrow$ / TP@1\% $\uparrow$ / TP@5\% $\uparrow$}}} \\ 
Algorithm & Train Domain & Test Domain & Pegasus & Pegasus-bigram & Parrot & Parrot-bigram \\ \midrule
 KGW & N/A &\texttt{BookSum} & 97.3 / \underline{89.7} / \underline{95.3} & 96.5 / 56.6 / 85.3 & 94.6 / 42.0 / 75.8 & 93.1 / 37.4 / 71.2 \\
 SIR & N/A &\texttt{BookSum} & 93.1 / 79.3 / 85.9 & 93.7 / 69.9 / 81.5 & 96.5 / \underline{72.9} / 85.1 & \underline{97.2} / \underline{76.5} / 88.0 \\
 \multirow{2}{*}{\shortmethodk} 
 & \texttt{RealNews} & \texttt{BookSum} & \underline{98.2} / 78.2 / 94.9 & \underline{97.3} / \underline{70.7} / \underline{93.8} & \underline{96.8} / 65.5 / \underline{90.9} & 96.4 / 61.9 / \underline{89.2} \\ 
 & \texttt{BookSum} & \texttt{BookSum} &\textbf{99.3 / 94.1 / 97.3} & \textbf{99.1 / 92.5 / 96.9} & \textbf{98.4 / 86.3 / 93.9} & \textbf{98.8 / 88.9 / 94.9} \\
 \midrule
\end{tabular}}
\caption{Ablation study on the detection robustness of \methodk (shown as \shortmethodk) to domain shifts. \textbf{Bold texts} mark the highest and \underline{underline texts} mark the second-highest result. In face of domain shifts, \methodk suffers a drop in performance yet is still able to retain some robustness over baselines we are comparing with.}
\label{tab:addk-res}
\end{table*}


\paragraph{Generation}
We use OPT-1.3B \citep{zhang2022opt} as our base autoregressive LM. To obtain robust sentence encoders specific to text domains for \methodk generations,
we fine-tune two versions of $\embd$, respectively on RealNews \citep{raffel2020exploring} and on BookSum \citep{kryscinski2021booksum} datasets (see \Sref{app:finetune} for specific procedure and parameter choices).

Following \citet{hou23semstamp} and  \citet{kirchenbauer2023watermark}, we sample at a temperature of 0.7 and a repetition penalty of 1.05, with 32 being the prompt length and 200 being the default generation length. Results with various lengths are included in Fig.~\ref{fig:length}. For \methodk, we perform $k$-means clustering on embeddings of sentences in 8k paragraphs, respectively on RealNews and BookSum. We keep $k=8$ and a valid region ratio $\gamma = 0.25$, which is consistent with the number of regions in \method, and we use a rejection margin $m = 0.035$.

\paragraph{Baselines}
Our baselines include popular watermarking algorithms \citet{kirchenbauer2023watermark}, \method, \textsc{Unigram-Watermark} \citep{zhao2023provable},  and the Semantic Invariant Robust (SIR) watermark in \citet{liu2023semantic}, implemented with their recommended setups. 


\subsection{Results}

\label{exp:results}
\paragraph{Detection} Detection results in Table \ref{tab:detection} show that \textbf{\methodk\ is more robust to paraphrase attacks than KGW \citep{kirchenbauer2023watermark} and \method}\ across Pegasus, Parrot, and GPT-3.5-Turbo paraphrasers and their bigram attack variants, as measured by AUC, TP@1\%, and TP@5\%. In particular, \methodk demonstrates considerable robustness against GPT-3.5, in which none of \method \ and KGW performed strongly. While \textsc{Unigram-Watermark} \cite{zhao2023provable} also demonstrates strong robustness against paraphrase, it has a critical vulnerability to reverse-engineering attacks. We discuss its vulnerability and experimental results in \Sref{app:add_exp_results}. The BERTScores of paraphrases are presented in Table \ref{tab:bertscore}.

\paragraph{Domain Shifts}
Since \methodk finetunes sentence-embedder from a specified text domain, we investigate the robustness of the fine-tuned sentence-embedder inputs from a different domain. In Table \ref{tab:addk-res}, we show that \methodk experiences a drop in robustness when using a cross-domain sentence-embedder. Nevertheless, \methodk\ is able to retain some robustness compared to KGW and SIR, staying especially resilient against Pegasus-bigram attacks.

\paragraph{Sampling Efficiency} 
\methodk not only demonstrates stronger paraphrastic robustness, but also \textbf{generates sentences with higher sampling efficiency.}
To produce the results on BookSum \citep{kryscinski2021booksum} in Table \ref{tab:detection}, \methodk samples 13.3 sentences on average to accept one valid sentence, which is 36.2\% less compared to the average 20.9 sentences sampled by \method. We analyze the reasons of candidate sentences for being rejected respectively by \methodk and \method, discovering that around 42.0\% and 80.7\% of the sentences are rejected due to the margin requirements. Since \methodk determines the cluster centroids by $k$-means clustering on the semantic structure of a given text domain, the embeddings of most candidate sentences generated in this text domain are closer to the centroids and away from the margins, and they are less likely to relocate to a blocked region after paraphrase. 

\paragraph{Quality} Table \ref{tab:quality} shows that
the perplexity, text diversity, and semantic diversity of both \method \ and \methodk generations are \textbf{on par with the base model without watermarking}, while KGW and SIR notably degrade perplexity. Qualitative examples of \methodk are presented in  
Figure \ref{fig:textexamples} and \ref{fig:textparaphraseexamples}. Compared to non-watermarked generation, \methodk convey the same level of coherence and contextual sensibility. The Ent-3 and Sem-Ent metrics also show that
\textbf{\methodk preserves token and semantic diversity of generation} compared to non-watermarked generation.
\begin{table}
    \centering
    \scalebox{0.75}{
    \begin{tabular}{rcccc}
        \toprule
        & \textbf{\textit{PPL}}$\downarrow$ & \textbf{\textit{Ent-3}}$\uparrow$ & \textbf{\textit{Sem-Ent}}$\uparrow$ \\ \midrule
        No watermark & 11.89 & 11.43 & 2.98 \\
        KGW & 14.92 & 11.32 & 2.95  \\
        SIR & 20.34 & 11.57 & 3.18 \\
        \method & 12.49 & 11.48 & 3.00 \\ 
        \methodk & 11.82 & 11.48 & 2.98 \\\bottomrule
    \end{tabular}
    }
    \caption{Quality evaluation of generations on BookSum. 
    $\uparrow$ and $\downarrow$ indicate the direction of preference (higher and lower). 
    \textbf{\methodk\ generation quality is on par with non-watermarked generations}.
    } 
    \label{tab:quality}
    \vspace{-4mm}
\end{table}

\begin{figure}[ht]
    \centering
    \includegraphics[scale=0.3]{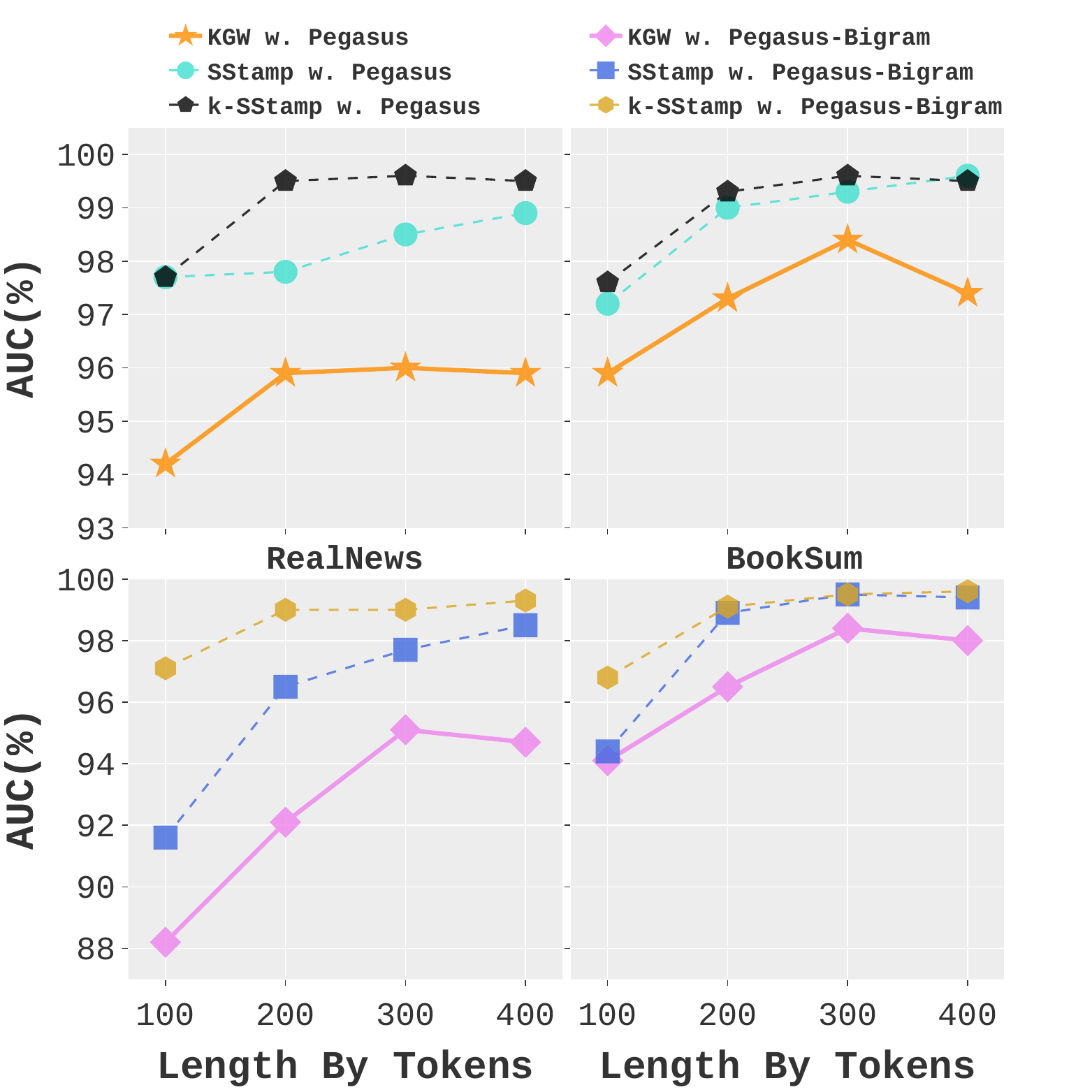}
    \vspace{-7mm}
    \caption{Detection results (AUC) under different generation lengths. \textbf{\methodk \ is more robust than \method\ and KGW across length 100-400 tokens in most cases.}} 
    \vspace{-5mm}
    \label{fig:length}
\end{figure}
\paragraph{Generation Length} As shown in Fig.~\ref{fig:length}, \methodk has higher AUC than \citet{kirchenbauer2023watermark} and than \method\ across most generation lengths by number of tokens.

\section{Conclusion}
\secvsbelow
We propose $k$-\method, a simple but effective enhancement of \method. To watermark generated sentences, \methodk maps embeddings of candidate sentences to a semantic space which is partitioned by $k$-means clustering, and only accept sampled sentences whose embeddings fall into a valid region. This variant greatly improves the paraphrastic robustness and sampling speed.

\secvsabove
\section*{Limitations}
\secvsbelow
\label{sec:discuss}


A core component of \methodk is performing $k$-means clustering on a particular text domain and partitioning the semantic space according to the semantic structure of the text domain. However, this requires specifying the text domain of generation to initialize \methodk. If the $k$-means clusters and the sentence embedder are not specific to the text domain, \methodk suffers from a minor drop in paraphrastic robustness (see Table \ref{tab:addk-res} for experimental results with \methodk using a sentence embedder trained on RealNews).

\secvsabove
\section*{Ethical Considerations}
\secvsbelow
The proliferation of large language models capable of generating realistic texts has drastically increased the need to detect machine-generated text. By proposing $k$-\method, we hope that practitioners will use this as a tool for governing model-generated texts. Although \methodk shows promising paraphrastic robustness, it is still not perfect for all kinds of attacks and thus should not be solely relied on in all scenarios. Finally, we hope this work motivates future research interests in not only semantic watermarking but also general adversarial-robust methods for AI governance.

\section*{Acknowledgement}
We would like to thank Brian Lu and following members of the Intelligence Amplification Lab: Yining Lu,  Nikil Sharma, Jiefu Ou, and Tianjian Li for their support and constructive feedback to this work. We are also grateful for the insightful advice from the broader JHU CLSP community and our anonymous reviewers and senior members at ACL.

\bibliography{custom,anthology}

\clearpage
\appendix
\section*{Supplemental Materials}

\section{Contrastive Learning and Sentence Encoder Fine-tuning}
\label{app:finetune}
To make sentence encoders robust to paraphrase, we fine-tune following the procedure in \citet{hou23semstamp} and \citet{wieting-etal-2022-paraphrastic}.

First, we paraphrase 8000 paragraphs from RealNews \citep{raffel2020exploring} and BookSum \citep{kryscinski2021booksum} using the Pegasus paraphraser \citep{zhang2020pegasus} through beam search with 25 beams. We then fine-tune two SBERT models\footnote{sentence-transformers/all-mpnet-base-v1} with an embedding dimension $h=768$ for 3 epochs with a learning rate of $4 \times 10^{-5}$, using the contrastive learning objective with a margin $\delta = 0.8$:
\begin{equation}
    \min_\theta \sum_i\max\Bigl\{\delta - f_\theta(s_i, t_i) + f_\theta(s_i, t'_i) , 0\Bigr\},
    \eqvs
\end{equation}
where $f_\theta$ measures the cosine similarity between sentence embeddings, $f_\theta(s, t) = \cos\bigl(M_\theta(s),M_\theta(t) \bigr)$, and $M_\theta$ is the sentence encoder parameterized by $\theta$ that is to be fine-tuned. 

\section{Algorithms}
\label{app:algo}
The algorithms of \methodk are presented in Algorithm \ref{alg:main-k}. 

\section{Watermark Detection}
\label{app:z-score}
The detection of both \method \ and \methodk follows the one-proportion $z$-test framework proposed by \citet{kirchenbauer2023watermark}. The $z$-test is performed on the number of green-list tokens in \citet{kirchenbauer2023watermark}, assuming the following null hypothesis:
\begin{nullhypothesis}
    \textit{A piece of text, T, is not generated (or written by human) knowing a watermarking green-list rule.}
\end{nullhypothesis}

The green-list token $z$-score is computed by:
\begin{equation}
\label{eq:z-score}
    z = \frac{N_G - \gamma N_T}{\sqrt{\gamma(1-\gamma)N_T}},
\end{equation}
where $N_G$ denotes the number of green tokens, $N_T$ refers to the total number of tokens contained in the given piece of text $T$, and $\gamma$ is a chosen ratio of green tokens. 

The $z$-test rejects the null hypothesis when the green-list token $z$-score exceeds a given threshold $M$. During the detection of each piece of text, the number of the green tokens is counted. A higher ratio of detected green tokens after normalization implies a higher $z$-score, meaning that the text is classified as machine-generated with more confidence. 

\citet{hou23semstamp} adapts this $z$-test to detect \method, according to the number of valid sentences rather than green-list tokens.
\
\begin{nullhypothesis}
    \label{null:semstampnull}
    \textit{ A piece of text, T, is not generated (or written by human) knowing a rule of valid and blocked partitions in the semantic space.}
\end{nullhypothesis}
 
\begin{equation}
    z = \frac{S_V - \gamma S_T}{\sqrt{\gamma(1-\gamma)S_T}},
\end{equation}
where $S_V$ refers to the number of valid \textit{sentences}, $\gamma$ is the ratio of valid sentences out of the total number of sentences $S_T$ in a piece of text $T$. 
To detect \method, the given piece of text, $T$, is first broken into sentences and the number of valid sentences $S_V$ is counted to calculate the $z$-score.
Likewise, the null hypothesis \ref{null:semstampnull} is rejected when the $z$-score exceeds a threshold M.

The detection procedure of \methodk is analogous to \method. We break a text into sentences and count the number of valid sentences to calculate the $z$-score, where only the determination of whether a sentence falls into a valid region is different. \methodk assigns the sentence generation to its closest cluster centroid and consider if the index of the cluster centroid belongs to a valid partition. See the full detection algorithm in Algorithm \ref{alg:detect}.

For a comprehensive evaluation of detection robustness, we consider a range of possible threholds $M_f \in [0, 6.0]$, where each $M_{f}$ is determined by a given false positive rate $r_f$, and consider machine-generated text as "positive" and human text as "negative" in a traditional classification setting. We estimate a false positive rate of corresponding $M_f$ by computing the misclassification rate of human text samples. We let $r_f = 0.01$ and $r_f = 0.05$ to respectively measure \textbf{\textit{TP@1\%}} and \textbf{\textit{TP@5\%}} metrics in Table \ref{tab:detection}. 

\begin{figure*}[t]
    \centering
    \fbox{
    \begin{minipage}{38.5em}
    \footnotesize
    \textbf{Prompt: }
    Pompey has a meeting with the pirates and they discuss the fact that Antony is in Egypt with Cleopatra.
    
    \leavevmode\newline
    \textbf{\method : }
    \textcolor{bettergreen}{The pirates want to ransom him, but they aren't sure how high the price would be. They have heard that Antony is a powerful monarch who can do whatever he wants. The pirates also say that he has a lot of money. Pompey says it wouldn't matter if he had 1,000 times as much money as Antony; he would still be a tyrant. The conversation turns to the bribes Antony was accused of taking. One pirate says that if Antony were a man of honor, he would have given up his rights to the Sea-Sir by now.
    } 
    \leavevmode\newline
    \textbf{Paraphrased \method: }
    \textcolor{bettergreen}{The pirates don't know how much they would get for him.} 
    \textcolor{red}{\ul{They have heard that Antony can do whatever he wants. He has a lot of money according to the pirates. }}\textcolor{bettergreen}{He would still be a tyrant if he had more money than Antony. Antony was accused of taking bribes. According to a pirate, Antony would have given up his rights to the Sea-Sir by now if he were a man of honor.
}
    \leavevmode\newline
    \leavevmode\newline
    \textbf{\methodk : }
    \textcolor{bettergreen}{They also discuss Caesar Augustus. The captain invites Pompey to share his thoughts on Caesar. He says that he doesn't like him very much. In fact, he thinks he's probably a bad leader. He goes on to say that Brutus should be his guest. Brutus replies that he can't because he's already got a guest for that matter - the Senate's spaghetti-spilling friend, Publius Cornelius.
    }
    \leavevmode\newline
    \textbf{Paraphrased \methodk: }
    \textcolor{bettergreen}{They talked about Caesar Augustus. Pompey was invited by the captain to share his thoughts on Caesar. He doesn't like him very much. He thinks he's a bad leader. He said that he should be his guest. Publius Cornelius is the Senate's spaghetti-spilling friend and he can't because he's already there.}
    \end{minipage}
    } 
    \caption{Examples of \methodk after being paraphrased by Pegasus Paraphraser \citep{zhang2020pegasus}. Green and plain sentences are detected, while red and underlined sentences are not. \textbf{\methodk generations are more robust to paraphrase, having a higher detection $z$-score than \method.}}
    \label{fig:textparaphraseexamples}
    \vspace{-2mm}
\end{figure*}

\begin{table*}[t]
\centering
\scalebox{0.88}{\begin{tabular}{@{}llllll@{}}
\toprule
 \multicolumn{6}{c}{\textit{\textbf{AUC / TP@1\% / TP@5\%}}} \\ 
Algorithm & Domain & Pegasus & Pegasus-bigram & Parrot & Parrot-bigram \\ \midrule
\multirow{2}{*}{\textsc{Unigram-Watermark}} & \texttt{RealNews} & 99.1 / 92.2 / 96.4 & 98.4 / 87.9 / 94.3 &98.9 / 82.7 / 94.0 &98.7 / 79.6 / 91.5 \\
 & \texttt{BookSum} & 99.4 / 96.4 / 99.0 & 99.7 / 91.6 / 98.2 & 99.5 / 91.6 / 97.7 & 99.6 / 87.8 / 97.2 \\\midrule
\end{tabular}}
\vspace{-2mm}
\caption{Detection results of \textsc{Unigram-Watermark} in \citet{zhao2023provable}}
\vspace{-6mm}
\label{app:unigram-res}
\end{table*}

\begin{table*}[t]
\small
\centering
\begin{tabular}{lcccccc}
\toprule
 & \multicolumn{3}{c}{\texttt{RealNews}} & \multicolumn{3}{c}{\texttt{BookSum}} \\ \cmidrule(l){2-7} 
\multicolumn{1}{l|}{\textbf{\textit{Algorithm}}$\downarrow$~\textbf{\textit{Paraphraser}}$\rightarrow$} & Pegasus & Parrot & \multicolumn{1}{c|}{GPT3.5} & Pegasus & Parrot & GPT3.5 \\ \midrule
\multicolumn{1}{c|}{KGW} & 71.0 / 66.6 & 57.1 / 58.4 & \multicolumn{1}{c|}{54.8 / 53.3} & 71.8 / 69.3 & 62.0 / 61.8 & 60.3 / 56.7 \\
\multicolumn{1}{c|}{\shortmethod} & 72.2 / 69.7 & 57.2 / 57.4 & \multicolumn{1}{c|}{55.1 / 53.8} & 73.0 / 71.3 & 64.4 / 67.1 & 55.4 / 50.0 \\ 
\multicolumn{1}{c|}{\shortmethodk} & 71.9 / 67.8 & 55.8 / 56.1 & \multicolumn{1}{c|}{54.8 / 53.3} & 73.5 / 71.5 & 64.2 / 67.1 & 35.7 / 33.4 \\ \bottomrule
\end{tabular}
\caption{BERTScore \citep{zhang2019bertscore} between original and paraphrased generations under different watermark algorithms and paraphrasers. All numbers are expressed in percentages. The first number in each entry is the result under regular sentence-level paraphrase attack in \citet{hou23semstamp}, while the second number is the result under the bigram paraphrase attack. \textbf{Compared to regular paraphrase attacks, bigram paraphrase attack only slightly corrupts the semantic similarity between paraphrased outputs and original generations.}}
\label{tab:bertscore}
\end{table*}

\section{Additional Experimental Results}
\label{app:add_exp_results}
Table \ref{app:unigram-res} shows the detection results of \textsc{Unigram-Watermark} \citep{zhao2023provable} against paraphrase attacks, demonstrating more robustness compared to \method \ and \methodk. However, \textsc{Unigram-Watermark} has the key vulnerability of being readily reverse-engineered by an adversary. Since \textsc{Unigram-Watermark} can be understood as a variant of the watermark in \citet{kirchenbauer2023watermark} but with only one fixed greenlist initialized at the onset of generation. An adversary can reverse-engineer this greenlist by brute-force submissions to the detection API of $|V|$ times, where each submission is repetition of a token $w_i$, $i\in \{1,...,|V|\}$ drawn without replacement from the vocabulary $V$ of the tokenizer. Therefore, upon each submission to the detection API, the adversary will be able to tell if the submitted token is in the greenlist or not. After $|V|$ times of submission, the entire greenlist can be reverse-engineered. On the other hand, such hacks are not applicable to \method\ and \methodk, since both algorithms do not fix the list of valid regions and blocked regions during generation. In summary, despite having strong robustness against various paraphrase attacks, \textsc{Unigram-Watermark} has a notable vulnerability that may limit its applicability in high-stake domains where adversaries can conduct reverse-engineering.

\paragraph{Computing Infrastruture and Budget}
We ran sampling and paraphrase attack jobs on 8 A40 and 4 A100 GPUs, taking up a total of around 200 GPU hours.




\begin{algorithm}[t]
\footnotesize

\end{algorithm}

\end{document}